# Application of a Fuzzy Programming Technique to Production Planning in the Textile Industry

*I. Elamvazuthi , T. Ganesan, P. Vasant
Universiti Technologi PETRONAS
Tronoh, Malaysia
*

J. F. Webb
Swinburne University of Technology Sarawak Campus,
Kuching, Sarawak, Malaysia

*Abstract*—Many engineering optimization problems can be considered as linear programming problems where all or some of the parameters involved are linguistic in nature. These can only be quantified using fuzzy sets. The aim of this paper is to solve a fuzzy linear programming problem in which the parameters involved are fuzzy quantities with logistic membership functions. To explore the applicability of the method a numerical example is considered to determine the monthly production planning quotas and profit of a home-textile group.

## I. Introduction

Many problems in science and engineering have been considered from the point of view optimization. As the environment is much influenced by the disturbance of social and economic factors, the optimization approach is not always the best. This is because, under such turbulent conditions, many problems are ill-defined. Therefore, a degree-of-satisfaction approach may be better than optimization. Here, we discuss how to deal with decision making problems that are described by fuzzy linear programming (FLP) models and formulated with elements of imprecision and uncertainty. More precisely, we will study FLP models in which the parameters are known only partially to some degree of precision.

Even though the information is incomplete, the model builder is able to provide realistic intervals for the parameters in these FLP models. We will demonstrate that the modeling complications can be handled with the help of some results which have been developed in fuzzy set theory. The FLP problem which we will be considering in this work is to find ways to handle fuzziness in the parameters. We will develop a FLP model in which the parameters are known with only some degree of precision. We will also show that the model can be parameterized in such a way that a satisfactory solution becomes a function of the membership values. The FLP model derived in this way is flexible and easy to handle computationally [1].

The first and most meaningful impetus towards the mathematical formalization of fuzziness was pioneered by Zadeh [2]. Its further development is in progress, with numerous attempts being made to explore the ability of fuzzy set theory to become a useful tool for adequate mathematical analysis of real-world problems [3]. The period of development of fuzzy theory from 1965 to 1977, is often referred to as the academic phase. The outcome was a rather small number of publications of a predominantly theoretical nature by a few contributors, mainly from the academic community. At this time, not much work in the area of fuzzy decision making was reported. The period from 1978 to 1988, has been called the transformation phase during which significant advances in fuzzy set theory were made and some real-life problems were solved. In this period, some important principles in fuzzy set theory and its applications were established. However, work on fuzzy decision making was not very active, in the area of engineering applications. Some earlier work on fuzzy decision making can be found in [4] and [5]. From 1989 to the present work on fuzzy techniques has boomed . In this period, many problems concerning applications in industry and business have been tackled successfully. In the early 1990s, fuzzy techniques were used to aid the solution of some soft computing problems. The aim of soft computing is to exploit, whenever possible, the tolerance for imprecision and uncertainty in order to achieve computational tractability, robustness, and low cost, by methods that produce approximate but acceptable solutions to complex problems which often have no precise solution.

Currently, fuzzy techniques are often applied in the field of decision making. Fuzzy methods have been developed in virtually all branches of decision making, including multi-objective, multi-person, and multi-stage decision making [6]. Apart from this, other research work connected to fuzzy decision making includes applications of fuzzy theory in management, business and operational research [7]. Some representative publications can be found in [8], [9], [10], [11] and [12].

Decision making is an important and much studied application of mathematical methods in various fields of human activity. In real-world situations, decisions are nearly always made on the basis of information which, at least in part, is fuzzy in nature. In some cases fuzzy information is used as an approximation to more precise information. This form of approximation can be convenient and sufficient for making good enough decisions in some situations. In other cases, fuzzy information is the only form of information available.

The first step in mathematically tackling a practical decision-making problem consists of formulating a suitable mathematical model of a system or situation. If we intend to make reasonably adequate mathematical models of situations that help practicing decision makers in searching for rational decisions, we should be able to introduce fuzziness into our models and to suggest means of processing fuzzy information.

In this paper a methodology to solve an FLP problem by using a logistic membership function is considered. The rest of the paper is organized as follows. In section 2, the basic fuzzy model is defined and this is followed by a numerical example in section 3. Section 4 provides the results and discussion, and finally, concluding remarks are made in section 5.

## II. THE MODEL

A conventional linear programming problem is defined by

Maximize $Cx$

Subject to $Ax \leq b, \quad x \geq 0.$ (1)

in which the components of a $1 \times n$ vector $C$, an $m \times n$ matrix $A$ and an $n \times 1$ vector $b$ are all crisp parameters and $x$ is an $n$-dimensional decision variable vector.

The system (1) may be redefined in a fuzzy environment with the following more elaborate structure:

Maximize $\sum_{j=1}^{n} \tilde{c}_j x_j$

Subject to

$$\sum_{j=1}^{n} \tilde{a}_{ij} x_j \leq b_i, \quad i = 1, 2 \cdots m \quad (2)$$

All fuzzy data $\tilde{c}_j \equiv \tilde{S}(c_j^a, c_j^b)$ and $\tilde{a}_{ij} \equiv \tilde{S}(a_{ij}^a, a_{ij}^b)$ are fuzzy variables with the following logistic membership functions [13],

$$\mu_{\tilde{c}_j} = \begin{cases} 1 & \text{if } c_j \leq c_j^a \\ \dfrac{B}{1 + Ce^{\alpha\left(\frac{c_j - c_j^a}{c_j^b - c_j^a}\right)}} & \text{if } c_j^a \leq c_j \leq c_j^b \\ 0 & \text{if } c_j \geq c_j^b \end{cases} \quad (3)$$

$$\mu_{\tilde{a}_{ij}} = \begin{cases} 1 & \text{if } a_{ij} \leq a_{ij}^a \\ \dfrac{B}{1 + Ce^{\alpha\left(\frac{a_{ij} - a_{ij}^a}{a_{ij}^b - a_{ij}^a}\right)}} & \text{if } a_{ij}^a \leq a_{ij} \leq a_{ij}^b \\ 0 & \text{if } a_{ij} \geq a_{ij}^b \end{cases} \quad (4)$$

## III. NUMERICAL EXAMPLE

In this example the profit for a unit of sheet sales is around 1.05 Euro; a unit of pillow case sales is around 0.3 Euro and a unit of quilt sales is around 1.8 Euro. The firm concerned would like to sell approximately 25.000 sheet units, 40.000 pillow case units and 10.000 units quilt units. The monthly working capacity and required process time for the production of sheets, pillow cases and quilts are given in Table 1 [14].

In view of this, let us determine monthly production planning details and profit for a home-textile group. $X_1$ presents the quantity of sheets that will be produced, $X_2$ presents the quantity of pillow cases and $X_3$ presents the quantity of quilts. The profit figures with logistic membership functions as given in Table I.

TABLE I. REQUIRED PROCESS TIME FOR SHEET, PILLOW CASE AND OF A QUILT [14]

| Departments | Required unit time(hour) | | | Working hours per month |
|---|---|---|---|---|
| | Sheet | Pillow case | Quilt | |
| Cutting | 0.0033 | 0.001 | 0.0033 | 208 |
| Sewing | 0.056 | 0.025 | 0.1 | 4368 |
| Pleating | 0.0067 | 0.004 | 0.017 | 520 |
| Packaging | 0.01 | 0.01 | 0.01 | 780 |

If we consider, around $1.05 \equiv \tilde{S}(1.02, 1.08)$, around $0.3 \equiv \tilde{S}(0.2, 0.4)$, and around $1.8 \equiv \tilde{S}(1.7, 2.0)$, then, the mathematical model of the above problem with fuzzy objective coefficients can be described as follows.

Maximize

$\tilde{S}(1.02,1.08)x_1 + \tilde{S}(0.2,0.4)x_2 + \tilde{S}(1.7,2.0)x_3$

subject to

$0.033\,x_1 + 0.01\,x_2 + 0.0033\,x_3 \leq 208$;

$0.056\,x_1 + 0.25\,x_2 + 0.1\,x_3 \leq 4368$;

$0.0067\,x_1 + 0.04\,x_2 + 0.17\,x_3 \leq 520$;

$0.1\,x_1 + 0.1\,x_2 + 0.01\,x_3 \leq 780$;

$x_1 \geq 25000$;

$x_2 \geq 40000$;

$x_3 \geq 10000$;

(5)

and we set $B = 1, C = .001, \varepsilon = 0.2$ and $d = 13.8$ [15].

The aspiration of the objective function is calculated by solving the following:

Maximize $1.08x_1 + 0.4x_2 + 2.0x_3$
subject to
$.0033x_1 + .001x_2 + .0033x_3 \leq 208$;
$.056x_1 + .025x_2 + .1x_3 \leq 4368$;
$.0067x_1 + .004x_2 + .017x_3 \leq 520$; (6)
$.01x_1 + .01x_2 + .01x_3 \leq 780$;
$x_1 \geq 25000$;
$x_2 \geq 40000$;
$x_3 \geq 10000$;

which gives the optimal value of the objective function as 67203.88 for $x_1 = 29126.21, x_2 = 35000.00$ and $x_3 = 10873.79$ [15].

With the help of the program LINGO version 10.0 we obtain the following results [15]:

$\lambda = 0.5323011, x_1 = 27766.99, x_2 = 40000.00,$
$x_3 = 10233.01, \eta = 0.4911863$

Therefore, to achieve maximum profit the home-textile group should plan for a monthly production of 27766.99 sheet units, 40000 pillow case units and 102333.01 quilt units. This plan gives an overall satisfaction of 0.5323011. The decision making method may be improved further by adopting a recursive iteration methodology.

IV. RESULTS AND DISCUSSION

The numerical example is solved by using a recursive method for various iterations. This was carried out using the C++ programming language on a personal computer with a dual core processor running at 2 GHz [16]–[17]. Fig. 1 shows the 3D outcome of the iterations with $M = 748$ for various alpha values with respect to the objective function $G$. The values of $\alpha_1$ and $\alpha_2$ vary from 0 to 1. The optimum values for the objective function as per Fig. 1 are 86,807.7 (maximum) and 86,755.4 (minimum).

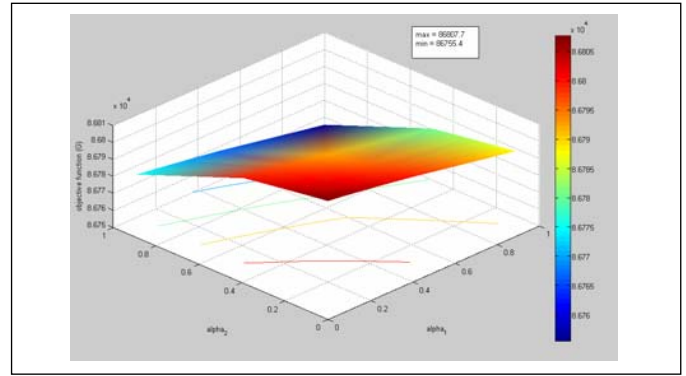

Figure 1. 3D plot for iterations M=748.

Fig. 2 shows the 3D outcome for $M = 749$ iterations and various alpha values with respect to $G$. The optimum values for the objective function as per this figure are 86,691.8 (maximum) and 86,639.5 (minimum).

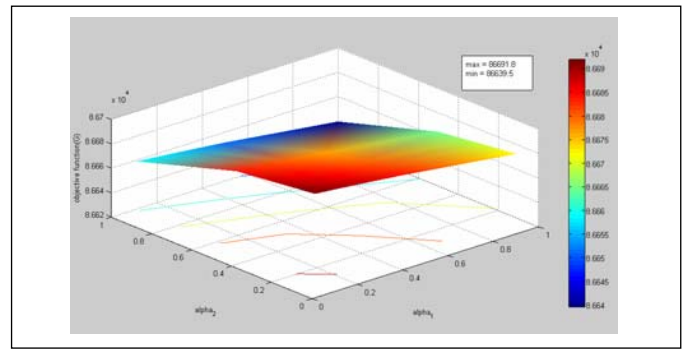

Figure 2. 3D plot for iterations $M = 749$.

Fig. 3 shows the 3D outcome for $M = 750$ iterations and various alpha values with respect to $G$. The optimum values for the objective function as per this figure are 86,576.2 (maximum) and 86,524.0 (minimum).

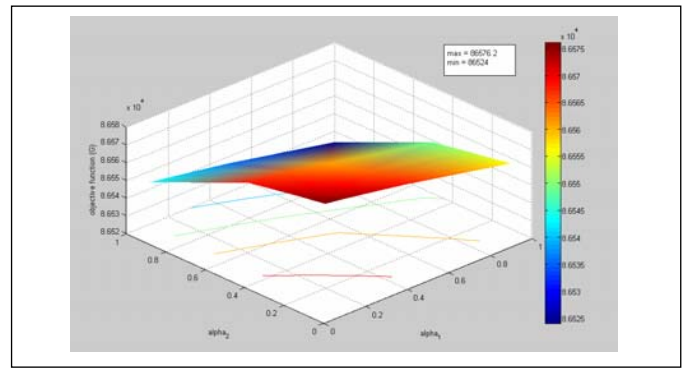

Figure 3. 3D plot for iterations M=750.

Fig. 4 shows the 3D outcome for $M = 751$ iterations and various alpha values with respect to $G$. The optimum value for the objective function as per this figure are 86,440.7 (maximum) and 86,408.0 (minimum).

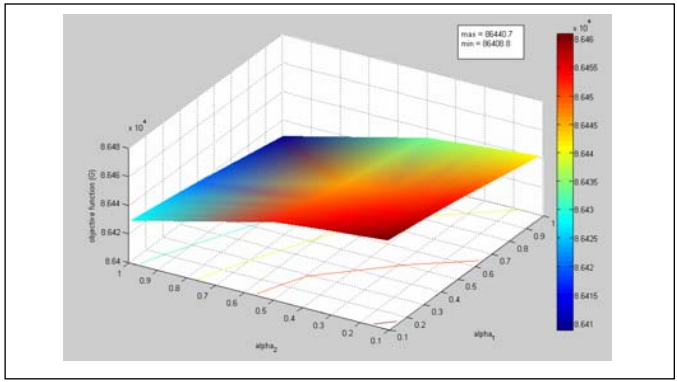

Figure 4. 3D plot for iterations M=751.

Fig. 5 shows the linear approximation for $G$ with respect to iterations 748 to 751. It can be seen that as the iterations are increased, the values of the objective function decrease. The percentage error is minimum at iteration, $M = 748$; however, after that it increases until it peaks at $M = 750$; thereafter, the percentage error decreases again to a level lower than that at $M = 748$. This shows that the maximum number of iterations that can be used for similar cases in the future can be limited to $M = 750$.

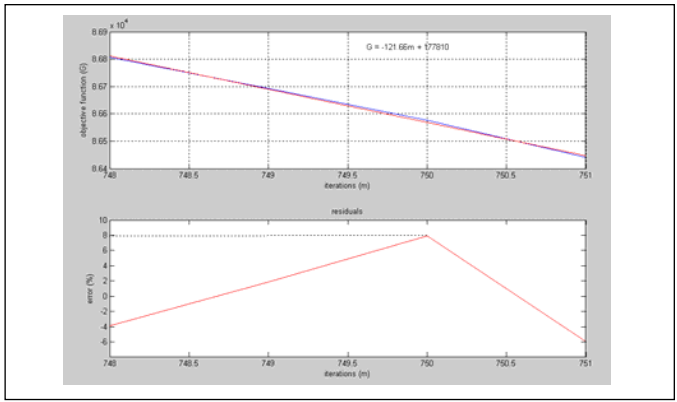

Figure 5. Objective Function (G) versus iterations

Figs. 6, 7 and 8 show the linear approximation for the decision variables $x_1$, $x_2$ and $x_3$ with respect to the number of iterations. It can be observed that $x_1$, $x_2$ and $x_3$ decrease as the iterations are increased from $M = 748$ to $M = 751$.

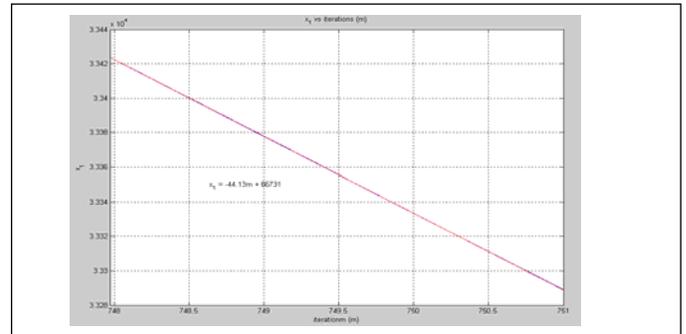

Figure 6. Decision variable, X1 versus M iterations.

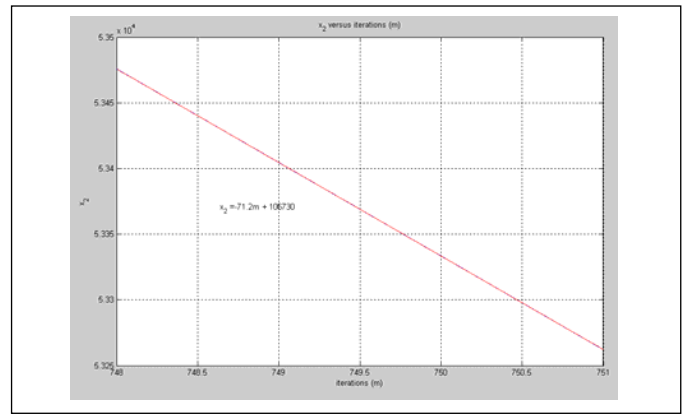

Figure 7. Decision variable, X2 versus M iterations.

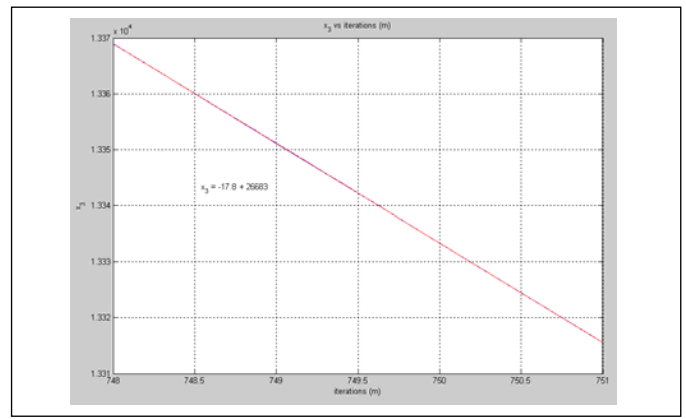

Figure 8. Decision variable, X2 versus M iterations.

Table II presents results that involve $\alpha_1$, $\alpha_2$ and $\alpha_3$ with $M = 748$ for $G$, $x_1$, $x_2$ and $x_3$. Other results for $M = 749$ to 751 are given in the appendix.

TABLE II ALPHA, OBJECTIVE FUNCTION
AND DECISION VARIABLES FOR M=748

| $\alpha_1$ | $\alpha_2$ | $\alpha_3$ | G | $x_1$ | $x_2$ | $x_3$ |
|---|---|---|---|---|---|---|
| 1 | 1 | *all | 86755.4 | 33422.5 | 53475.9 | 13369 |
| 1 | 0.5 | all | 86780.3 | 33422.5 | 53475.9 | 13369 |
| 0.5 | 1 | all | 86767.5 | 33422.5 | 53475.9 | 13369 |
| 0.5 | 0.5 | all | 86792.4 | 33422.5 | 53475.9 | 13369 |
| 0.3333 | 1 | all | 86770.9 | 33422.5 | 53475.9 | 13369 |
| 0.3333 | 0.5 | all | 86795.9 | 33422.5 | 53475.9 | 13369 |
| 0.25 | 1 | all | 86772.5 | 33422.5 | 53475.9 | 13369 |
| 0.25 | 0.5 | all | 86797.5 | 33422.5 | 53475.9 | 13369 |
| 0.2 | 1 | all | 86773.5 | 33422.5 | 53475.9 | 13369 |
| 0.2 | 0.5 | all | 86798.4 | 33422.5 | 53475.9 | 13369 |
| 0.1667 | 1 | all | 86774.1 | 33422.5 | 53475.9 | 13369 |
| 0.1667 | 0.5 | all | 86799.0 | 33422.5 | 53475.9 | 13369 |
| 0.1429 | 1 | all | 86774.5 | 33422.5 | 53475.9 | 13369 |
| 0.1429 | 0.5 | all | 86799.4 | 33422.5 | 53475.9 | 13369 |
| 0.125 | 1 | all | 86774.8 | 33422.5 | 53475.9 | 13369 |
| 0.125 | 0.5 | all | 86799.8 | 33422.5 | 53475.9 | 13369 |
| 0.1111 | 1 | all | 86775.1 | 33422.5 | 53475.9 | 13369 |
| 0.1111 | 0.5 | all | 86800.0 | 33422.5 | 53475.9 | 13369 |

Note: *all $\in (0, 1)$

Table III summarizes the result for $M = 748$ to 751 for $x_1$, $x_2$ and $x_3$ with maximum and minimum values of $G$. The overall maximum value for $G$ is 86807.7 at $M = 748$ and the overall minimum value is 86408.0 at $M = 751$.

TABLE III SUMMARY OF ITERATIONS, DECISION VARIABLES
AND OBJECTIVE FUNCTION

| M | $x_1$ | $x_2$ | $x_3$ | G (max) | G(min) |
|---|---|---|---|---|---|
| 748 | 33422.5 | 53475.9 | 13369 | 86807.7 | 86755.4 |
| 749 | 33377.8 | 53404.5 | 13351.1 | 86691.8 | 86639.5 |
| 750 | 33333.3 | 53333.3 | 13333.3 | 86576.2 | 86524.0 |
| 751 | 33288.9 | 53262.3 | 13315.6 | 86440.7 | 86408.0 |

Table IV compares the best objective function and decision variables $x_1$, $x_2$ and $x_3$ of the proposed method with previous work by other researchers.

TABLE IV COMPARATIVE ANALYSIS

| Method | The Best Objective Function | Decision Variables | | |
|---|---|---|---|---|
| | | $x_1$ | $x_2$ | $x_3$ |
| Irfan [14] | 64390.999 | 33825.16 | 40000.00 | 9374.760 |
| Atanu [15] | 66454.369 | 27766.99 | 40000.00 | 10233.01 |
| Proposed Method | 86807.700 | 33422.50 | 53475.90 | 13369.00 |

From Table IV, the optimum value for the objective function using the proposed method outweighs the results obtained in [14] and [15]. It can be deduced that the recursive iteration method proposed here is an efficient and effective way to solve our example fuzzy problem of production planning in the textile industry.

V. CONCLUSION

This paper has discussed the use of fuzzy linear programming for solving a production planning problem in the textile industry. It can be concluded that the recursive method introduced is a promising method for solving such problems. The modified s-curve membership function provides various uncertainty levels which are very useful in the decision making process. In this paper, only a single s-curve membership function was considered. In the future, various other membership functions will be considered. Apart from providing an optimum solution for the objective functions, the proposed method ensures high productivity. In this regard, there is a good opportunity for developing an interactive self-organized decision making method by using hybrid soft computing techniques.

ACKNOWLEDGMENT

The authors would like to thank Universiti Teknologi PETRONAS and Swinburne University of Technology Sarawak Campus for supporting this work.

APPENDIX

TABLE V ALPHA, OBJECTIVE FUNCTION AND DECISION VARIABLES FOR M=749

| $\alpha_1$ | $\alpha_2$ | $\alpha_3$ | G | $x_1$ | $x_2$ | $x_3$ |
|---|---|---|---|---|---|---|
| 1 | 1 | *all | 86639.5 | 33377.8 | 53404.5 | 13351.1 |
| 1 | 0.5 | all | 86664.4 | 33377.8 | 53404.5 | 13351.1 |
| 0.5 | 1 | all | 86651.7 | 33377.8 | 53404.5 | 13351.1 |
| 0.5 | 0.5 | all | 86676.6 | 33377.8 | 53404.5 | 13351.1 |
| 0.3333 | 1 | all | 86655.1 | 33377.8 | 53404.5 | 13351.1 |
| 0.3333 | 0.5 | all | 86680.0 | 33377.8 | 53404.5 | 13351.1 |
| 0.25 | 1 | all | 86656.7 | 33377.8 | 53404.5 | 13351.1 |
| 0.25 | 0.5 | all | 86681.6 | 33377.8 | 53404.5 | 13351.1 |
| 0.2 | 1 | all | 86657.6 | 33377.8 | 53404.5 | 13351.1 |
| 0.2 | 0.5 | all | 86682.5 | 33377.8 | 53404.5 | 13351.1 |
| 0.1667 | 1 | all | 86658.2 | 33377.8 | 53404.5 | 13351.1 |
| 0.1667 | 0.5 | all | 86683.1 | 33377.8 | 53404.5 | 13351.1 |
| 0.1429 | 1 | all | 86658.7 | 33377.8 | 53404.5 | 13351.1 |
| 0.1429 | 0.5 | all | 86683.5 | 33377.8 | 53404.5 | 13351.1 |
| 0.125 | 1 | all | 86659.0 | 33377.8 | 53404.5 | 13351.1 |
| 0.125 | 0.5 | all | 86683.9 | 33377.8 | 53404.5 | 13351.1 |
| 0.1111 | 1 | all | 86659.2 | 33377.8 | 53404.5 | 13351.1 |
| 0.1111 | 0.5 | all | 86684.4 | 33377.8 | 53404.5 | 13351.1 |

Note: *all $\in$ (0, 1) , M= no. of iterations

TABLE VI ALPHA, OBJECTIVE FUNCTION AND DECISION VARIABLES FOR M=750

| $\alpha_1$ | $\alpha_2$ | $\alpha_3$ | G | $x_1$ | $x_2$ | $x_3$ |
|---|---|---|---|---|---|---|
| 1 | 1 | *all | 86524 | 33333.3 | 53333.3 | 13333.3 |
| 1 | 0.5 | all | 86548.9 | 33333.3 | 53333.3 | 13333.3 |
| 0.5 | 1 | all | 86536.1 | 33333.3 | 53333.3 | 13333.3 |
| 0.5 | 0.5 | all | 86561 | 33333.3 | 53333.3 | 13333.3 |
| 0.3333 | 1 | all | 86539.5 | 33333.3 | 53333.3 | 13333.3 |
| 0.3333 | 0.5 | all | 86564.4 | 33333.3 | 53333.3 | 13333.3 |
| 0.25 | 1 | all | 86541.1 | 33333.3 | 53333.3 | 13333.3 |
| 0.25 | 0.5 | all | 86566 | 33333.3 | 53333.3 | 13333.3 |
| 0.2 | 1 | all | 86542.1 | 33333.3 | 53333.3 | 13333.3 |
| 0.2 | 0.5 | all | 86566.9 | 33333.3 | 53333.3 | 13333.3 |
| 0.1667 | 1 | all | 86542.7 | 33333.3 | 53333.3 | 13333.3 |
| 0.1667 | 0.5 | all | 86567.5 | 33333.3 | 53333.3 | 13333.3 |
| 0.1429 | 1 | all | 86543.1 | 33333.3 | 53333.3 | 13333.3 |
| 0.1429 | 0.5 | all | 86568 | 33333.3 | 53333.3 | 13333.3 |
| 0.125 | 1 | all | 86543.4 | 33333.3 | 53333.3 | 13333.3 |
| 0.125 | 0.5 | all | 86568.3 | 33333.3 | 53333.3 | 13333.3 |
| 0.1111 | 1 | all | 86543.7 | 33333.3 | 53333.3 | 13333.3 |
| 0.1111 | 0.5 | all | 86568.5 | 33333.3 | 53333.3 | 13333.3 |

Note: *all $\in$ (0, 1)

TABLE VII ALPHA, OBJECTIVE FUNCTION AND DECISION VARIABLES FOR M=751

| $\alpha_1$ | $\alpha_2$ | $\alpha_3$ | G | $x_1$ | $x_2$ | $x_3$ |
|---|---|---|---|---|---|---|
| 1 | 1 | *all | 86408.8 | 33288.9 | 53262.3 | 13315.6 |
| 1 | 0.5 | all | 86433.6 | 33288.9 | 53262.3 | 13315.6 |
| 0.5 | 1 | all | 86420.9 | 33288.9 | 53262.3 | 13315.6 |
| 0.5 | 0.5 | all | 86445.7 | 33288.9 | 53262.3 | 13315.6 |
| 0.3333 | 1 | all | 86424.3 | 33288.9 | 53262.3 | 13315.6 |
| 0.3333 | 0.5 | all | 86449.1 | 33288.9 | 53262.3 | 13315.6 |
| 0.25 | 1 | all | 86425.9 | 33288.9 | 53262.3 | 13315.6 |
| 0.25 | 0.5 | all | 86450.7 | 33288.9 | 53262.3 | 13315.6 |
| 0.2 | 1 | all | 86426.9 | 33288.9 | 53262.3 | 13315.6 |
| 0.2 | 0.5 | all | 86451.7 | 33288.9 | 53262.3 | 13315.6 |
| 0.1667 | 1 | all | 86427.5 | 33288.9 | 53262.3 | 13315.6 |
| 0.1667 | 0.5 | all | 86452.3 | 33288.9 | 53262.3 | 13315.6 |
| 0.1429 | 1 | all | 86427.9 | 33288.9 | 53262.3 | 13315.6 |
| 0.1429 | 0.5 | all | 86452.7 | 33288.9 | 53262.3 | 13315.6 |
| 0.125 | 1 | all | 86428.2 | 33288.9 | 53262.3 | 13315.6 |
| 0.125 | 0.5 | all | 86453 | 33288.9 | 53262.3 | 13315.6 |
| 0.1111 | 1 | all | 86428.4 | 33288.9 | 53262.3 | 13315.6 |
| 0.1111 | 0.5 | all | 86453.3 | 33288.9 | 53262.3 | 13315.6 |

Note: *all $\in$ (0, 1)

AUTHOR PROFILES

I. Elamvazuthi is a lecturer in the Department of Electrical and Electronic Engineering, Universiti Teknologi PETRONAS (UTP), Malaysia. His research interests include Control Systems, Mechatronics and Robotics.

T. Ganesan is currently a Graduate Assistant with the Department of Mechanical Engineering, Universiti Teknologi PETRONAS (UTP), Malaysia, pursuing a Masters Degree. He has a Bachelor's Degree in Mechanical Engineering from the same university. He specializes in Computational Fluid Mechanics.

P. Vasant is a lecturer in the Department of Fundamental and Applied Sciences, Universiti Teknologi PETRONAS (UTP), Malaysia. His research interests are Soft Computing and Computational Intelligence.

J. F. Webb is a lecturer at Swinburne University of Technology, Sarawak Campus, Kuching, Sarawak, Malaysia. He specializes in Computational Methods, Nano-Physics and Ferroelectric Materials.